\documentclass[letterpaper, 10 pt, conference]{ieeeconf}  
\usepackage[T1]{fontenc}

\IEEEoverridecommandlockouts                              
\usepackage{graphicx} 
\usepackage{amsfonts,amssymb}
\usepackage{stfloats}
\usepackage{booktabs}
\usepackage{multirow}
\usepackage{float}
\title{\LARGE \bf
MSTF: Multiscale Transformer for Incomplete Trajectory Prediction}

\author{Zhanwen Liu$^{1}$, Chao Li$^{1}$, Nan Yang$^{1}$, Yang Wang$^{1*}$, Jiaqi Ma$^{2}$, Guangliang Cheng$^{3}$ and Xiangmo Zhao$^{1}$
\thanks{This research was funded by the National Natural Science Foundation of China (General Program) [No. 52172302], the Two-chain Integration Key Special Project of Shaanxi Provincial Department of Science and Technology - Enterprise-Institute Joint Key special Project [2023-LL-QY-24], and Shannxi Province Traffic science and Technology Program [21-02X]}
\thanks{* Corresponding author}
\thanks{$^{1}$Zhanwen Liu, Chao Li, Nan Yang, Yang Wang, Yang Wang, Xiangmo Zhao are with the Department of Information Engineering, Chang’an University, Xi’an, Shaanxi 710018, PR China. {\tt\small zwliu@chd.edu.cn, lichao971204@foxmail.com, 2022024001@chd.edu.cn, ywang120@ustc.edu.cn, xmzhao@chd.edu.cn}}%
\thanks{$^{2}$Jiaqi Ma is with the UCLA Mobility Lab and FHWA Center of Excellence on New Mobility and Automated Vehicles, University of California, Los Angeles (UCLA), CA
90095 USA. {\tt\small jiaqima@ucla.edu}}%
\thanks{$^{3}$Guangliang Cheng is with the Department of Computer Science, University of Liverpool, L69 3BX Liverpool, U.K. {\tt\small  guangliangcheng2014@gmail.com}}%
}

\begin{document}

\maketitle
\thispagestyle{empty}
\pagestyle{empty}

\begin{abstract}
Motion forecasting plays a pivotal role in autonomous driving systems, enabling vehicles to execute collision warnings and rational local-path planning based on predictions of the surrounding vehicles. However, prevalent methods often assume complete observed trajectories, neglecting the potential impact of missing values induced by object occlusion, scope limitation, and sensor failures. Such oversights inevitably compromise the accuracy of trajectory predictions. To tackle this challenge, we propose an end-to-end framework, termed Multiscale Transformer (MSTF), meticulously crafted for incomplete trajectory prediction. MSTF integrates a Multiscale Attention Head (MAH) and an Information Increment-based Pattern Adaptive (IIPA) module. Specifically, the MAH component concurrently captures multiscale motion representation of trajectory sequence from various temporal granularities, utilizing a multi-head attention mechanism. This approach facilitates the modeling of global dependencies in motion across different scales, thereby mitigating the adverse effects of missing values. Additionally, the IIPA module adaptively extracts continuity representation of motion across time steps by analyzing missing patterns in the data. The continuity representation delineates motion trend at a higher level, guiding MSTF to generate predictions consistent with motion continuity. We evaluate our proposed MSTF model using two large-scale real-world datasets. Experimental results demonstrate that MSTF surpasses state-of-the-art (SOTA) models in the task of incomplete trajectory prediction, showcasing its efficacy in addressing the challenges posed by missing values in motion forecasting for autonomous driving systems.
\end{abstract}

\section{INTRODUCTION}
Predicting the future trajectory of vehicles is an essential task for autonomous driving systems. Autonomous vehicles (AVs) are empowered to conduct more reasonable local-path planning and collision warning based on the trajectory predictions of surrounding vehicles, which greatly improves the efficiency and safety of AVs in complex dynamic traffic systems. Based on sensory information derived from roadside or onboard sensing systems, such as vehicle location and road topology \cite{ref1,ref2,ref3,ref4,ref5,ref6,ref7,ref8,ref9}, existing methods typically perform temporal inference of the future trajectory by various well-designed models \cite{ref10,ref11,ref12,ref13,ref14,ref15}. Traditional approaches involve rasterizing the traffic scene and employing RNN-based models to capture temporal dependencies, yielding promising results in simple highway scenarios \cite{ref16,ref17,ref18}. However, the intricate road topology of urban traffic scenes poses inherent challenge to the rasterization paradigm. In response, graph-based models \cite{ref19,ref20} have been introduced for flexible prediction within non-Euclidean space, which notably outperforms RNN-based models, particularly in scenarios with complex road networks and dynamic urban traffic scenes. The emergence of Transformer \cite{ref21} has further advanced trajectory prediction. Transformer-based models \cite{ref22} establish direct links for inputs, allowing them to capture long-term dependencies in trajectory, and advancing the state of the art in long-term trajectory prediction.

\begin{figure}[thpb]
      \centering
      \includegraphics[scale=0.3]{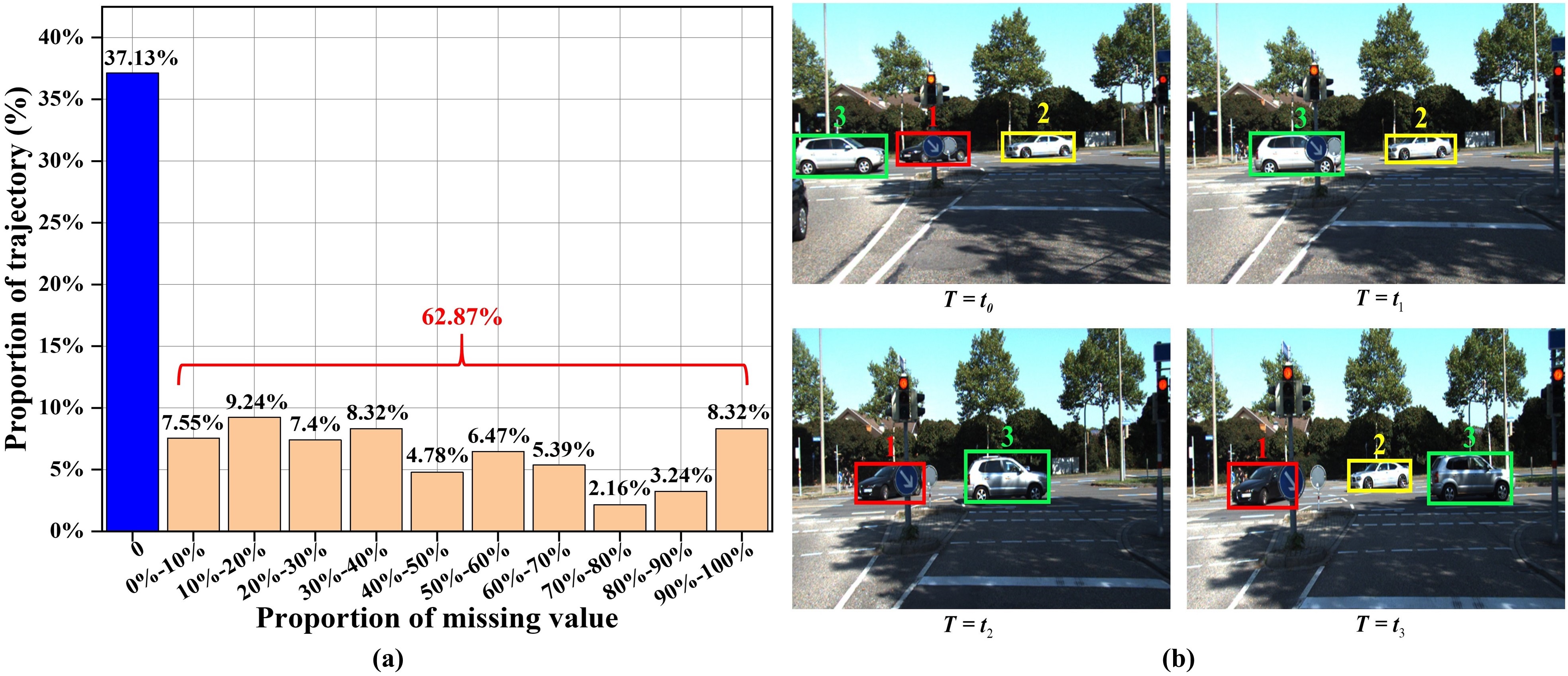}
      \caption{(a) lists the distribution of missing percentage of trajectory, showing that most of the trajectory have varying proportions of missing values. In the case shown in (b), vehicle 1 and vehicle 2 are occluded by vehicle 3 at time ${t_1}$ and ${t_2}$, respectively, resulting in missing values for their trajectory.}
      \label{fig1}
\end{figure}

However, existing methods often assume that the observed trajectory of the vehicle is entirely complete while \textbf{\textit{ignoring the potential for missing values}} caused by object occlusion, sensor failures, and sensing scope limitation. To elucidate this concern, we statistically analyze the missing values of the trajectory using the multi-object tracking dataset from KITTI \cite{ref23}, as shown in Fig. \ref{fig1}. In Fig. \ref{fig1} (a), distribution of missing percentages in trajectories is presented, revealing that a mere 37.13\% of vehicle trajectory samples exhibit completeness, while 62.87\% of the samples lack trajectory values at specific intervals. Notably, the missing percentages are dispersed randomly across the entire range of (0\%, 100\%). Fig. \ref{fig1} (b) provides an illustrative example of an occlusion case. The missing values disrupt temporal dependence of the trajectory sequence, and predicting the future trajectory of vehicles under such circumstance undoubtedly hinders the performance and negatively influences the behavior understanding of vehicles.

Although various recent methods have been proposed to solve the problem of missing values by imputation \cite{ref24,ref25,ref26}, most are autoregressive models that impute current missing values based on previous time steps, making them highly susceptible to compounding errors, especially in long-term temporal modeling. Additionally, widely used benchmarks are not tailored for the precise demands of vehicle trajectory prediction \cite{ref27,ref28}. More importantly, the two-stage incomplete trajectory prediction scheme that incorporates an imputation task brings extra parameters and computation burden, which hinders the lightweight and timeliness of autonomous driving systems.

In this paper, we present an end-to-end framework for incomplete trajectory prediction, Multiscale Transformer (MSTF). Specifically, we design a novel Multiscale Attention Head (MAH) leveraging the padding mask mechanism in the vanilla Transformer, and MAH observes the incomplete trajectory from different temporal granularities parallelly to extract multiscale motion representation. Meanwhile, we propose an Information Increment-based Pattern Adaptive (IIPA) module, capable of adaptively computing the information increment of various time steps utilizing the trajectory missing pattern (number and location of missing values, etc.), and model the motion continuity representation across time steps based on the information increment. The critical idea behind our method is that the motion representation at different scales may skip certain missing values, and the negative impact of missing values can be alleviated by using the multi-scale motion representation to predict the current value from different temporal granularities. Furthermore, the continuity representation reflects the overall trend of motion and is insensitive to the missing patterns of trajectory. It sacrifices part of the detailed information but can guide MSTF to output predictions that are consistent with motion consistency. It loses part of the detailed information but can guide MSTF to output predictions that are consistent with motion consistency. The main contributions of our work can be summarized as follows:

\begin{itemize}
\item We statistically analyze the problem of missing values of trajectory in real traffic scenarios and devise an end-to-end framework, MSFT, for incomplete trajectory prediction. The MAH is designed to capture multi-scale motion representations of vehicles from different temporal granularities, mitigating the negative impact of missing values on vehicle trajectory prediction.
\item We propose a novel IIPA module that is able to adaptively compute information increments at different time steps using trajectory missing patterns, and then model missing pattern insensitive continuity representation across time steps to guide MSTF to output prediction that is consistent with motion consistency.
\item Through comparative experiments on both highway and urban scene datasets, MSTF consistently demonstrates superior performance compared to the existing SOTA methods.
\end{itemize}

\begin{figure*}[thpb]
      \centering
      \includegraphics[scale=0.43]{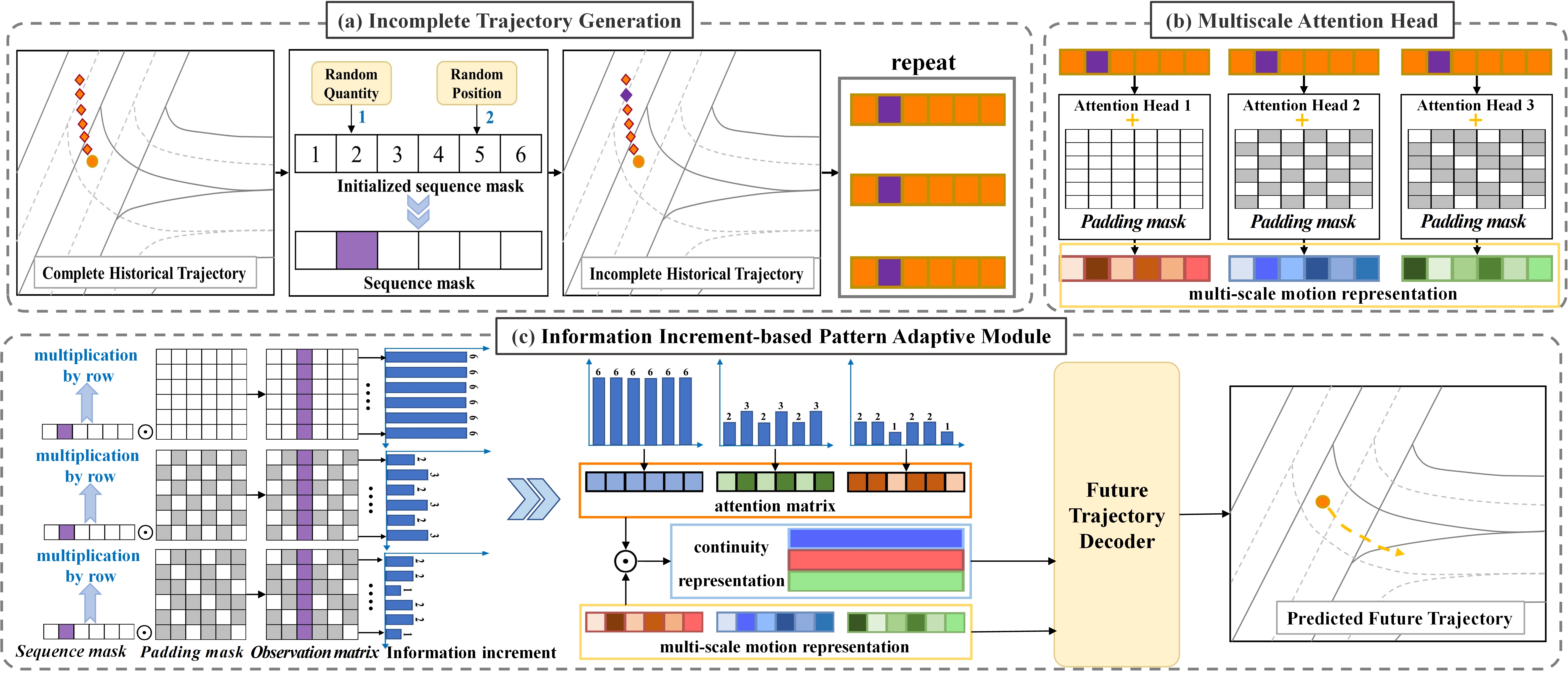}
      \caption{Illustration of the proposed MSTF framework. (a) Generate the sequence mask matrix with randomly distributed number and position of masks, which is used to mask the complete trajectory provided by the public dataset to obtain incomplete trajectory. (b) Construct multiscale attention head by predefined padding mask matrix with different temporal granularities for extracting multi-scale motion representation. (c) Perform information incremental analysis based on the sequence mask matrix and the padding mask matrix to obtain continuity representation across time steps. The future trajectory decoder outputs the future trajectory based on the multi-scale motion representation and continuity representation.}
      \label{fig2}
\end{figure*}

\section{RELATED WORK}
\subsection{Trajectory Prediction}
The objective of trajectory prediction is to predict the future positions of vehicles conditioned on their observations through various well-designed models. As a typical representative of RNN models, Social-LSTM \cite{ref16} innovatively embeds vehicle features by rasterizing traffic scenes for interaction extraction, and then sequentially decodes future trajectory through the recursive work mechanism of LSTM. Following this, other LSTM-based methods have been proposed \cite{ref17,ref18}. For special traffic scenes such as roundabouts and intersections, graph-based methods are proposed to adapt to complex road topology, facilitating the vehicle trajectory prediction in non-Euclidean space \cite{ref19,ref29}. Recently, Transformer-based models \cite{ref22} have been applied to this task to establish direct links for inputs via an attentional mechanism, allowing the models to capture long-term dependency of the trajectory. However, these methods assume that vehicle observations are entirely complete, which is too strong an assumption to satisfy in practice. Existing methods are not applicable to the prediction of incomplete trajectory whose temporal dependency is disrupted by missing values.

\subsection{Trajectory Imputation}
Some statistical imputation techniques substitute missing values with mean or median values \cite{ref30}. Alternative methods also adopt linear fitting \cite{ref31}, k-nearest neighbors \cite{ref32}, and expectation-maximization algorithm \cite{ref33}. One of the inherent limitations of such methods is that they use rigid prior, which hinders the generalization ability. In contrast, deep learning-based frameworks perform imputation more flexibly. For instance, some RNN-based models \cite{ref34} estimate missing values in sequences through deep autoregression, and generative models \cite{ref35} reconstruct incomplete sequences through GANs or VAEs. Nevertheless, the two-stage incomplete trajectory prediction framework of imputation followed by prediction brings extra parameters and computation burden, which hinders the lightweight and timeliness of autonomous driving systems. Therefore, we designed a novel framework called MSTF based on Transformer \cite{ref21}, which enables end-to-end incomplete trajectory prediction by extracting multi-scale motion representation and continuity representation.

\section{METHODS}
\subsection{Problem Definition}
Due to the manual annotation, trajectory data provided by the existing large public datasets \cite{ref36,ref37} is complete, and the incomplete trajectory is unavailable. To address this limitation, we generate incomplete trajectory by randomly concealing portions of the complete data. Specifically, consider a set of complete vehicle observations $X = \left\{ {{x^{t + 1}},{x^{t + 2}},...,{x^{t + {T_h}}}} \right\}$ over time step $t + 1$ to $t + {T_h}$, which is provided by public dataset, where ${x^t}\in\mathbb{R}^2$ represents the 2D coordinates of vehicle at time step $t$. To model the missing of vehicle observations due to occlusion, sensor failure, etc., we define a sequence mask matrix ${M_s} = \left\{ {m_s^{t + 1},m_s^{t + 2},...,m_s^{t + {T_h}}} \right\}$ valued in $\left\{ {0,1} \right\}$. The variable $m_s^t$ is assigned a value of 0 if the observation is missing at time step $t$ and 1 otherwise, and the quantity and positions of absent observations are generated in a fully random manner. Following this setting, the generated incomplete trajectory can be expressed as:
\begin{equation}
{X_{miss}} = X \odot {M_s}
\end{equation}
where ${X_{miss}}$ is the randomly masked incomplete trajectory, and the training, validation and testing of the model are performed based on the incomplete trajectory.

The goal of the incomplete trajectory prediction task is to predict the vehicle trajectory $\hat Y = \left\{ {{{\hat y}^{t + {T_h} + 1}},{{\hat y}^{t + {T_h} + 2}},...,{{\hat y}^{t + {T_h} + {T_f}}}} \right\}$ within the future time step $t + {T_h} + 1$ to $t + {T_h} + {T_f}$, conditioned on its incomplete observations over time step $t + 1$ to $t + {T_h}$, where ${T_h}$ and ${T_f}$ are the observation and prediction horizons, respectively.

\subsection{Model Framework}
Fig. \ref{fig2} provides a high-level depiction of our proposed framework. Firstly, the sequence mask matrix is obtained by randomly generating the number and distribution positions of masks, which is used to mask the complete trajectory provided by the public dataset to obtain the incomplete trajectory. Then, the incomplete trajectory is repeated and fed to multiple attention heads with different temporal granularities to extract the multi-scale motion representation. Finally, based on the sequence mask matrix and predefined padding mask matrix, the information incremental analysis at different temporal scales is performed for the weighted aggregation of the multi-scale motion representation across time steps to obtain continuity representation. Combining the detailed motion information expressed in the multi-scale motion representation with the overall trend of the motion reflected in the continuity representation, the future trajectory decoder outputs prediction for incomplete trajectory.

\subsection{Multiscale Attention Head}
The core of trajectory prediction lies in effectively modeling the temporal dependency between historical trajectory points, while the presence of missing values disrupts the dependency between adjacent time steps. We argue that RNN encoders (e.g., LSTM-based or GRU-based encoders) that serially process data using a recursive mechanism will undoubtedly rely more on local dependency between adjacent time steps, which makes their performance more susceptible to the negative impact of missing values. On the contrary, the Transformer processes the sequence of trajectory in parallel and is able to establish direct links for all values of the sequence with the help of an attention mechanism, so that each value in the sequence can directly aggregate information from all the remaining values to obtain global dependency, which alleviates the negative impact of some missing values to a certain extent. Consequently, designing the encoder based on Transformer in our work is a natural decision.

\begin{figure}[thpb]
      \centering
      \includegraphics[scale=0.53]{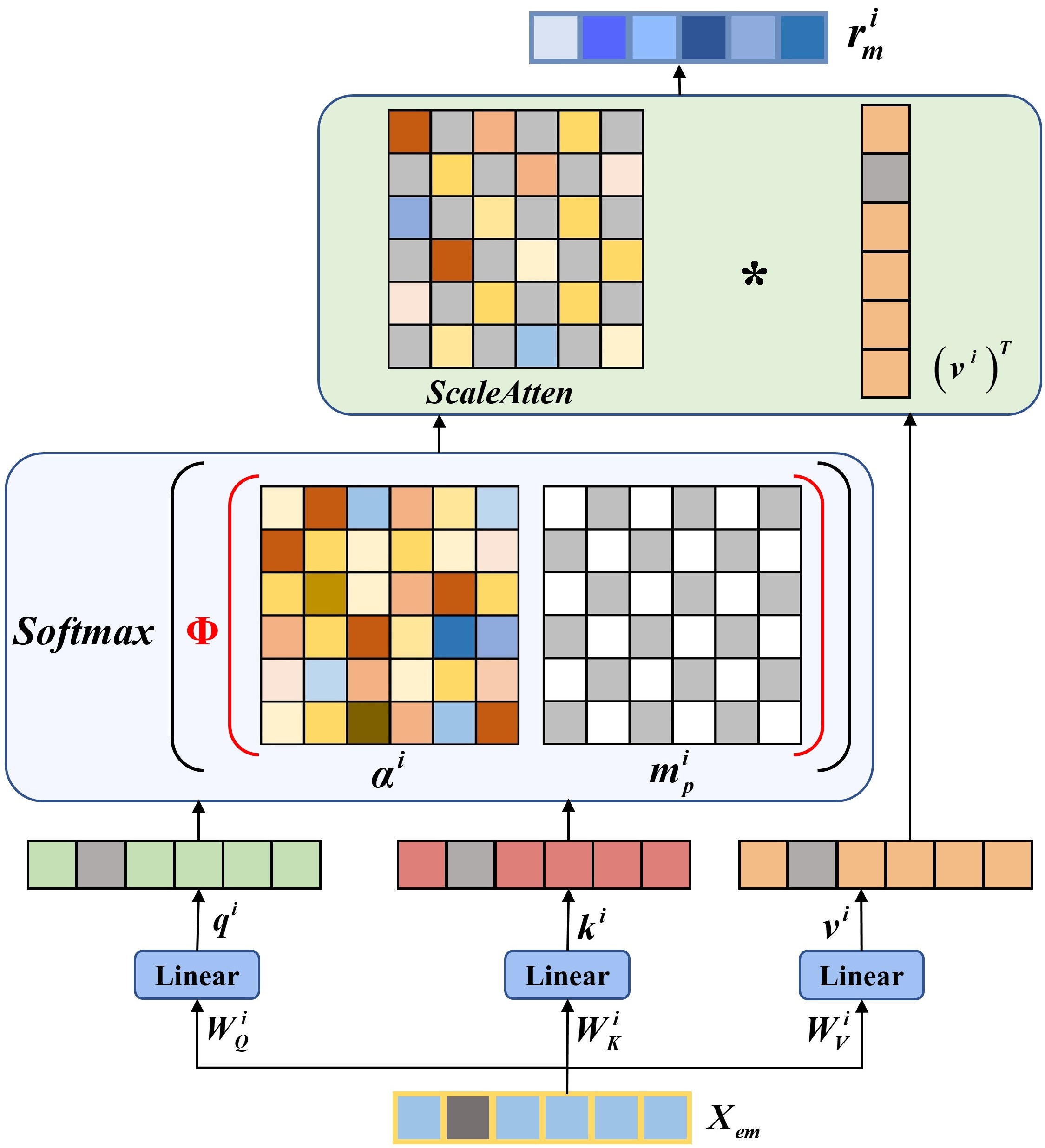}
      \caption{The computation process for the attention head $i$. Padding mask is the core that determines the temporal scale of the attention head, and different attention heads are identical except for padding mask. In this example, $m_p^i$ is the padding mask matrix for $i = 2$, where the gray squares are 0 and the white ones are 1.}
      \label{fig3}
\end{figure}

Specifically, we first compute the query vector $Q = \left\{ {{q^1},{q^2},...,{q^n}} \right\}$, the key vector $K = \left\{ {{k^1},{k^2},...,{k^n}} \right\}$, and the value vector $V = \left\{ {{v^1},{v^2},...,{v^n}} \right\}$ for the $n$ attention heads based on the incomplete input.
\begin{equation}
\begin{array}{l}
{X_{em}} = \beta ({X_{miss}}) + Pos\\
{q^i} = {\varphi _Q}\left( {{X_{em}},W_Q^i} \right)\\
{k^i} = {\varphi _K}\left( {{X_{em}},W_K^i} \right)\\
{v^i} = {\varphi _V}\left( {{X_{em}},W_V^i} \right)
\end{array}
\end{equation}
where $\beta$ is used to extend two-dimensional coordinates to higher dimension to improve feature representation, which is achieved through MLP in our work. Following Transformer \cite{ref21}, positional encoding $Pos$ is adopted to the model to distinguish the order of input sequence. $W_Q^i$, $W_K^i$, and $W_V^i$ are the learnable parameter matrices for corresponding transformation ${\varphi _Q}$, ${\varphi _K}$, and ${\varphi _V}$.

For different attention heads, the padding mask matrix ${M_p} \in {\mathbb{R}^{n \times len \times len}}$ with different temporal granularities is designed:
\begin{equation}
{M_p} = \left\{ {m_p^1,m_p^2,...,m_p^n} \right\}
\end{equation}

where ${m_p^i} \in {\mathbb{R}^{len \times len}}$ is the padding mask matrix of the attention head $i$, $n$ is the number of attention heads, and $len$ represents the length of input sequence. 

The value of element $\delta _{a,b}^i$ in row $a$ and column $b$ of matrix $m_p^i$ can be further formulated as:
\begin{equation}
\delta _{a,b}^i = \left\{ \begin{array}{l}
1,{\rm{ }}\frac{{a - b}}{i} \in \mathbb{Z}\\
0,{\rm{ }}Others
\end{array} \right.{\rm{        }}a,b \in \left\{ {1,2,...,len} \right\}
\end{equation}
where $\mathbb{Z}$ denotes the set of integers. For intuitive illustration,  we visualize the padding mask matrix $m_p^i$ for $i = 2$ in Fig. \ref{fig3}.

Based on the padding mask matrix, multiple attention heads extract the multi-scale motion representation ${R_m} = \left\{ {r_m^1,r_m^2,...,r_m^n} \right\}$ of the vehicle in parallel with different temporal granularities.
\begin{equation}
\begin{array}{l}
{\alpha ^i} = {q^i}{\left( {{k^i}} \right)^T}\\
ScaleAtten\left( {{\alpha ^i},m_p^i} \right) = soft\max \left( {\frac{{\Phi \left( {{\alpha ^i},m_p^i} \right)}}{{\sqrt {d_k^i} }}} \right)\\
r_m^i = ScaleAtten\left( {{\alpha ^i},m_p^i} \right)*{\left( {{v^i}} \right)^T}
\end{array}
\end{equation}
where $r_m^i$ is the motion representation extracted by the attention head $i$. $\Phi$ is a mapping function, which is used to map the value in ${\alpha ^i}$ at the position corresponding to the value 0 in $m_p^i$ to negative infinity. $d_k^i$ represents the dimension of key vector ${k^i}$, and the number of attention heads $n = 5$ in practice. The complete computation process of attention head $i$ is shown in Fig. \ref{fig3}.

\subsection{Information Increment-based Pattern Adaptive Module}
The absence of trajectory points hinders the model from adequately capturing the temporal dependency within the trajectory sequence. This challenge is particularly pronounced for RNN-based models, as they struggle to effectively capture the local dependency between consecutive time steps. The randomly generated missing patterns (the number of missing values and their distributed locations) also make the encoded feature of the same trajectory sample vary randomly with the missing patterns, which poses a great challenge to accurately decode the future trajectory of the vehicle. We argue that humans are not constrained by the locality of the sequence when facing the problem of incomplete trajectory prediction. Instead, they analyze the continuity of motion from a higher-level perspective. The continuity representation cannot encapsulate the detailed information of vehicle motion, but it aptly reflects the overall trend of motion across time steps and is insensitive to the missing patterns of trajectory, which is conducive to constraining the model to output the prediction consistent with the motion trend. Given the aforementioned analysis, we propose an Information Increment-based Pattern Adaptive (IIPA) module to extract the continuity representation.

Formally, based on the randomly generated sequence mask matrix ${M_s}$ and the predefined padding mask matrix ${M_p} = \left\{ {m_p^1,m_p^2,...,m_p^n} \right\}$, the observation matrix ${M_{obs}} = \left\{ {m_{obs}^1,m_{obs}^2,...,m_{obs}^n} \right\}$ is computed:
\begin{equation}
m_{obs}^i = \Lambda \left( {{M_s},m_p^i} \right)
\end{equation}
where sequence mask matrix ${M_s}$ represents the missing pattern of trajectory. The padding mask matrix $m_p^i \in {\mathbb{R}^{len \times len}}$ and observation matrix $m_{obs}^i$ reflect the scale of the observation and the observable values when the temporal granularity is $i$, respectively. $\Lambda$ denotes that ${M_s}$ and $m_p^i$ are multiplied by their corresponding elements row by row.

Then, based on the observation matrix $m_{obs}^i$, we statistically analyze the increment of information ${\Omega ^i} = \left[ {\sigma _1^i,\sigma _2^i,...,\sigma _{len}^i} \right]$ of the sequence at the temporal granularity $i$.
\begin{equation}
\begin{array}{l}
\mu _{j,l}^i \in \left\{ {0,1} \right\}\\
\sigma _j^i = \sum\limits_{l = 1}^{len} {\mu _{j,l}^i} 
\end{array}
\end{equation}
where $\mu _{j,l}^i$ is the value of the observation matrix $m_{obs}^i$ in row $j$ and column $l$. $\mu _{j,l}^i = 0$ indicates that $l-th$ trajectory point is missing or not within the observational scope of $j-th$ trajectory point at temporal granularity $i$, which renders the $j-th$ trajectory point incapable of aggregating information from the $l-th$ trajectory point through the attention mechanism; Otherwise, $l-th$ trajectory point is available for $j-th$ trajectory point. $\sigma _j^i$ is the information increment of $j-th$ trajectory point in the trajectory sequence when the temporal granularity is $i$. 

The multiscale attention head establishes a direct link for each value in the input sequence with the help of attention mechanism, which enables them to directly aggregate global information. Consequently, the feature of each trajectory point in the multi-scale motion representation can reflect the overall trend of the motion to a certain extent, only that the trajectory points at different locations observe the motion trend from different perspectives. Therefore, multi-scale motion representation is aggregated across time steps to synthesize different perspectives to obtain robust continuity representation. Specifically, considering the different impact of missing values on trajectory points at different locations, we compute the attention weights across time steps based on the information increment and give greater weight to the features of the trajectory points that are less affected by missing values, and finally obtain the continuity representation ${R_c} = \left\{ {r_c^1,r_c^2,...,r_c^n} \right\}$ that is insensitive to missing patterns.
\begin{equation}
\begin{array}{l}
a_j^i = \frac{{\exp \left( {\sigma _j^i} \right)}}{{\sum\nolimits_{l = 1}^{len} {\exp \left( {\sigma _l^i} \right)} }}\\
AcrossAtten\left( {{\Omega ^i}} \right) = \left\{ {a_1^i,a_2^i,...,a_{len}^i} \right\}\\
r_c^i = AcrossAtten\left( {{\Omega ^i}} \right) \times {\left( {r_m^i} \right)^T}
\end{array}
\end{equation}
where $r_{c}^i$ represents the continuity representation at temporal granularity $i$.

Finally, based on the multi-scale motion representation ${R_m}$ and continuity representation ${R_c}$, the future trajectory decoder combines the motion detail information with the overall trend of the motion to output the future prediction.
\begin{equation}
\begin{array}{l}
R = {\rm{AGG}}\left( {{R_m},{R_c}} \right)\\
\hat Y = \mathcal{P}\left( R \right)
\end{array}
\end{equation}
where ${\rm{AGG}}$ stands for data fusion, which is realized by concatenation in our work. $\mathcal{P}$ is $LSTM$, which is used as the future trajectory decoder.

\section{EXPERIMENTS}
\subsection{Datasets}
\begin{figure}[thpb]
      \centering
      \includegraphics[scale=0.52]{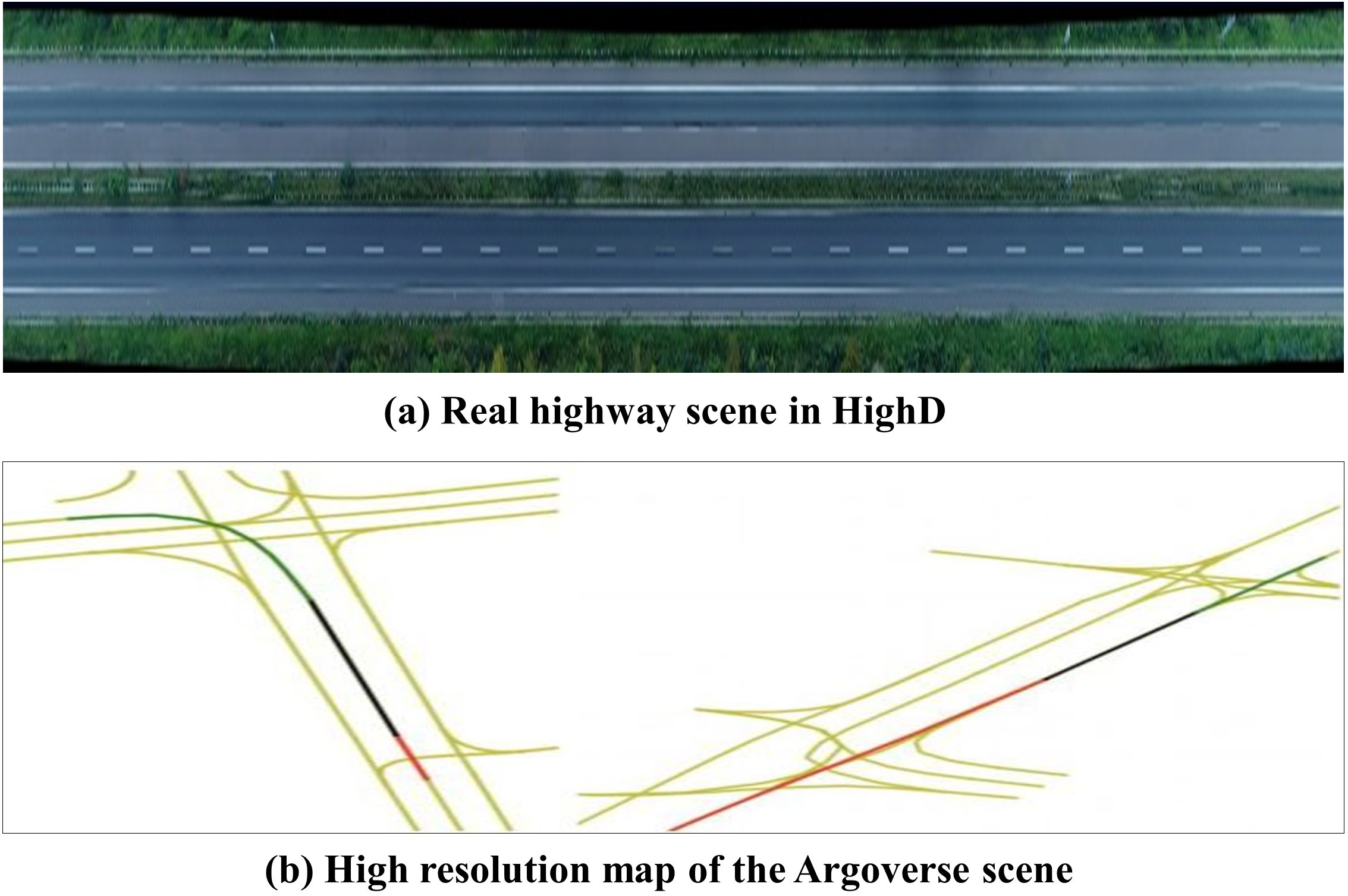}
      \caption{(a) shows a real highway scene where the HighD dataset was collected. (b) HD map data provided by the Argoverse dataset, showing the complex road topology where the data was collected.}
      \label{fig4}
\end{figure}

Considering the difference of vehicle behavior in highway traffic scenarios and urban traffic scenarios, we validate the validity of the proposed model in different traffic scenarios by using the HighD dataset \cite{ref36} and the Argoverse dataset \cite{ref37}, respectively. The HighD dataset was collected from the German highway as shown in Fig. \ref{fig4} (a), where vehicles in the traffic scenario travel faster, but with simple traffic behaviors such as acceleration, deceleration, and lane changing only. The data is recorded at 25Hz from six different locations on Germany highway from the aerial perspective using a drone. It is composed of 60 recordings over areas of 400$~$420 meters span, with a mileage of 45,000 km, and more than 110, 000 vehicles are contained.

Argoverse is a motion prediction benchmark that collects more than 30K data based on the onboard sensing system in urban traffic scenarios as shown in Fig. \ref{fig4} (b), where vehicles are slow but have complex traffic behaviors such as left or right turns. Each scenario is a 5-second sequence sampled at 10 Hz, and the task is to predict the position of the vehicle in the next 3 seconds based on its historical trajectory over 2 seconds. The sequences are split into training, validation, and test sets, which have 205942, 39472, and 78143 sequences respectively. In our work, we only use historical vehicle trajectory for prediction and do not use map data such as rasterized drivable area maps and ground height maps provided by the benchmark.

\subsection{Evaluation Metrics}
To facilitate the performance comparison, we follow previous works \cite{ref16,ref17,ref19,ref38} and use different evaluation metrics on HighD dataset and Argoverse dataset. In the comparison based on the HighD dataset, we use the root mean square error ($RMSE$) to evaluate the performance of the model at different prediction horizons. In the comparison based on the Argoverse dataset, the average displacement error (${\rm{ADE}}$) and final displacement error (${\rm{FDE}}$) are adopted to evaluate models. In order to make a fair comparison with our proposed model, we only use the single prediction of the existing models for the evaluation, although they give multiple possible predictions for the same sample.
\begin{equation}
\begin{array}{l}
\begin{array}{l}
RMSE = \sqrt {\frac{1}{m}\sum\limits_{i = 1}^m {\frac{1}{{{T_f}}}\sum\limits_{t = {T_h} + 1}^{{T_h} + {T_f}} {{{\left( {\hat y_i^t - y_i^t} \right)}^2}} } } \\
ADE = \frac{1}{{m{T_f}}}\sum\limits_{i = 1}^m {\sum\limits_{t = {T_h} + 1}^{{T_h} + {T_f}} {\sqrt {{{\left( {\hat y_i^t - y_i^t} \right)}^2}} } } \\
FDE = \frac{1}{m}\sum\limits_{i = 1}^m {\sqrt {{{\left( {\hat y_i^{{T_h} + {T_f}} - y_i^{{T_h} + {T_f}}} \right)}^2}} } 
\end{array}
\end{array}
\end{equation}
where $m$ is the number of samples. $\hat y_i^t$ and $y_i^t$ are the predicted and true positions of the sample $i$ at time $t$, which are 2D coordinates.

\subsection{Implementation Details}
We implement MSTF in PyTorch and train on 1 NVIDIA GeForce RTX 3090 with a batch size of 128. The MSTF has four layers, each consisting of five attention heads with different temporal scales, all possessing a hidden layer dimension of 128. We use Adam to train the model for 200 epochs, and set the initial learning rate to $1 \times {10^{ - 4}}$. We keep the same setting for both datasets.

\subsection{Results} 
To assess the prediction performance of existing models on incomplete trajectory, we only consider SOTA models with available code for comparison. The parameter settings of the following comparison models are set to default values, and only the trajectory is randomly masked to get incomplete input.
\begin{itemize}
\item Vanilla-Transformer (V-TF): The Vanilla-Transformer model with exactly the same structure as our proposed MSTF (number of attention heads, number of layers, dimensions of hidden states) is used as an ablation experiment to compare with the MSTF to demonstrate the validity of our proposed modules.
\item CS-LSTM \cite{ref16}: The method introduces convolutional operation into the social pooling layer to capture the inter-vehicle interaction while retaining the spatial information between vehicles. The output of CS-LSTM is a binary Gaussian distribution parameter.
\item PiP \cite{ref17}: The model couples trajectory prediction as well as the planning of the target vehicle by conditioning on multiple candidate trajectory of the target vehicle, and the facilitation between planning and prediction enables the model to achieve accurate predictions in highway traffic scenarios.
\item LaneGCN \cite{ref19}:The method proposes a fusion network consisting of four types of interaction based on graph convolution to model actor-lane, lane-lane, lane-actor and actor-actor interaction, and achieves accurate multimodal trajectory prediction with the help of this structured map representation and actor-map interaction.
\item HLS \cite{ref38}: The method introduces a hierarchical latent structure into VAE-based forecasting model. Based on the assumption that the trajectory distribution can be approximated as a mixture of simple distributions (or modes), the method employs low-level and high-level latent variables to model each mode of the mixture and the weights for the modes, respectively, which achieves promising prediction performance in complex urban traffic scenarios.
\end{itemize}

To evaluate the performance of the models across varying degrees of missing data, we delineate three distinct missing rate intervals, which are  $\left( {0\% ,30\% } \right]$,  $\left( {30\% ,60\% } \right]$, and  $\left( {60\% ,90\% } \right]$. Within these three intervals, the number and locations of missing trajectory points are randomly generated.

Table. \ref{table1} shows the comparison results of the models based on the HighD dataset. In general, our proposed MSTF achieves optimal prediction accuracy in all experiment settings. Through the comparison of V-TF with CS-LSTM and PiP, it can be seen that although the performance of V-TF is not the best among the three in short-term prediction (1 s), the average performance improvement of V-TF in long-term prediction (2s-4s) reaches 67.49\%, 67.89\% and 64.63\% in the three missing rate intervals, respectively. This indicates that the missing data disrupts the local dependency of adjacent time steps, which makes the performance of PiP and CS-LSTM degrade significantly, while V-TF can model the global dependency using the attention mechanism, which enables it to maintain better prediction performance even when the missing rate becomes large. Furthermore, the average performance improvement of MSTF over V-TF is 20.23\%, 12.86\%, and 11.39\% in the three missing rate intervals, respectively. Since the structure (number of attention heads, number of layers, dimension of hidden state) of V-TF and MSTF are identical, the comparison reveals that the performance improvement of MSTF can be attributed to the MAH and IIPA modules we proposed, rather than a mere expansion of model parameters.


\begin{table}[H]
\renewcommand{\arraystretch}{1.2}
\caption{The results of comparative experiment on HighD dataset.}
\label{table1}
\centering
\begin{tabular}{c|c|c|c|c|c}
\hline
Missing Rate&Horizons&CS-LSTM&PiP&V-TF&MSTF\\
\cline{1-6}
\multirow{5}{*}{ $\left( {0\% ,30\% } \right]$}&1 s&0.29&0.54&0.31&\textbf{0.19} \\
\cline{2-6}
                         &2 s&0.76&1.21&0.39&\textbf{0.30} \\
\cline{2-6}
                         &3 s&1.47&2.10&0.57&\textbf{0.47} \\
\cline{2-6}
                         &4 s&2.32&3.22&0.78&\textbf{0.69} \\
\cline{2-6}
                         &5 s&3.64&4.58&1.07&\textbf{0.96} \\
\cline{1-6}

\multirow{5}{*}{ $\left( {30\% ,60\% } \right]$}&1 s&0.31&0.62&0.33&\textbf{0.26} \\
\cline{2-6}
                         &2 s&0.81&1.35&0.41&\textbf{0.35} \\
\cline{2-6}
                         &3 s&1.52&2.29&0.59&\textbf{0.52} \\
\cline{2-6}
                         &4 s&2.40&3.44&0.82&\textbf{0.75} \\
\cline{2-6}
                         &5 s&3.72&4.82&1.12&\textbf{1.03} \\
\cline{1-6}

\multirow{5}{*}{ $\left( {60\% ,90\% } \right]$}&1 s&0.39&0.73&0.39&\textbf{0.34} \\
\cline{2-6}
                         &2 s&0.97&1.58&0.51&\textbf{0.45} \\
\cline{2-6}
                         &3 s&1.72&2.61&0.74&\textbf{0.66} \\
\cline{2-6}
                         &4 s&2.65&3.84&1.03&\textbf{0.92} \\
\cline{2-6}
                         &5 s&3.96&5.27&1.38&\textbf{1.23} \\

\hline
\end{tabular}
\end{table}

\begin{figure*}[thpb]
      \centering
      \includegraphics[scale=0.179]{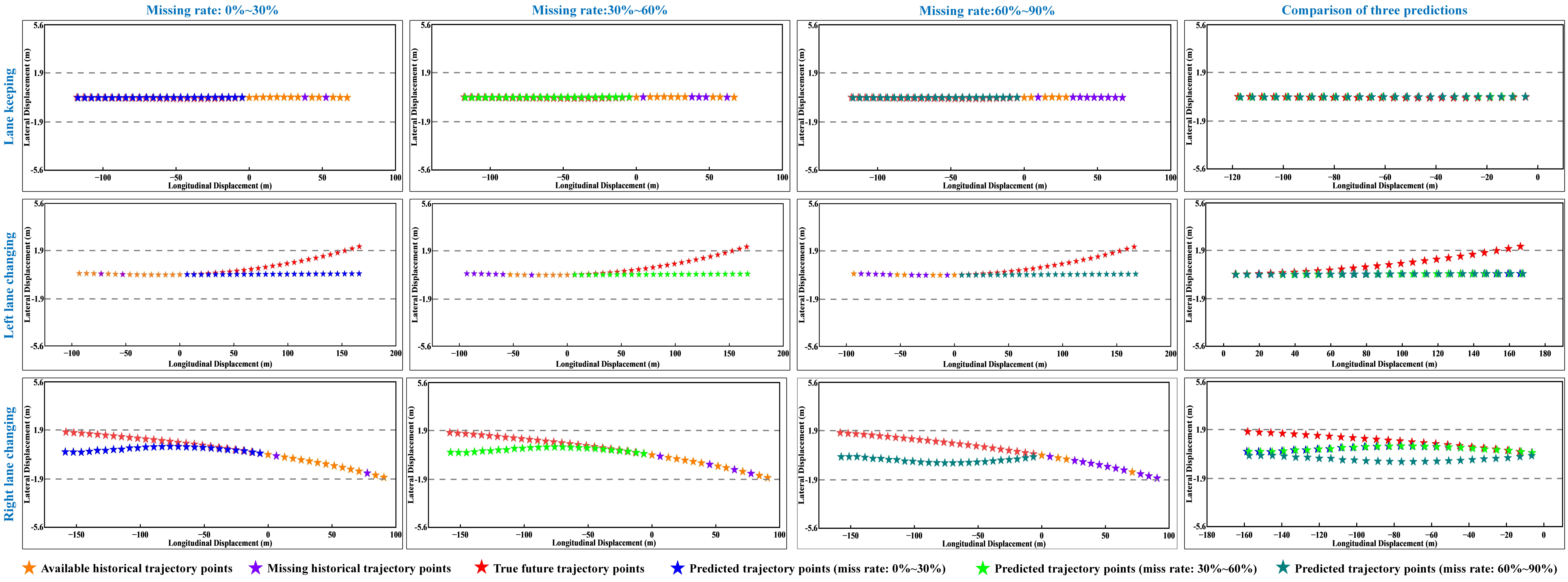}
      \caption{Visualization of predictions for three different maneuvers at different missing rate intervals.}
      \label{fig5}
\end{figure*}

The quantitative experimental results on Argoverse dataset are summarized in Table \ref{table2}. HLS significantly outperforms LaneGCN in complete trajectory prediction, but obtains the largest prediction error in comparison experiments for incomplete trajectory prediction, with even worse performance than V-TF, which illustrates that existing models designed for the complete trajectory prediction task cannot be flexibly transfered to incomplete trajectory prediction task. However, compared to LaneGCN, the MSTF designed for incomplete trajectory prediction only achieves 18.53\% and 11.05\% performance improvement for ADE and FDE when the missing interval is $\left( {60\% ,90\% } \right]$, while the prediction performance is worse than LaneGCN in all other experimental settings. We argue that the complex road topology makes the vehicle trajectory in the Argoverse dataset exhibit a high degree of nonlinearity, which affects the extraction of continuity representation by IIPA and ultimately limits the prediction performance of MSTF. In contrast, LaneGCN fully utilizes the high-definition map information and achieves reconstruction of missing information with the help of four well-designed interaction modules, which enables it to achieve excellent performance in the task of incomplete trajectories in complex scenes.

\begin{table}[H]
\renewcommand{\arraystretch}{1.2}
\caption{The results of comparative experiment on Argoverse dataset.}
\label{table2}
\centering
\begin{tabular}{c|c|c|c|c|c}
\hline
Missing Rate&Metric&LaneGCN&HLS&V-TF&MSTF\\
\cline{1-6}
\multirow{2}{*}{ $\left( {0\% ,30\% } \right]$}&ADE&\textbf{1.49}&2.20&2.00&1.91 \\
\cline{2-6}
                         &FDE&\textbf{3.23}&4.56&4.33&4.26 \\
\cline{1-6}
\multirow{2}{*}{ $\left( {30\% ,60\% } \right]$}&ADE&\textbf{1.78}&2.40&2.05&1.95 \\
\cline{2-6}
                         &FDE&\textbf{3.76}&4.96&4.45&4.34 \\
\cline{1-6}
\multirow{2}{*}{ $\left( {60\% ,90\% } \right]$}&ADE&2.59&2.85&2.25&\textbf{2.11} \\
\cline{2-6}
                         &FDE&5.25&5.66&4.83&\textbf{4.67} \\
\hline
\end{tabular}
\end{table}

To visually show the prediction effect of MSTF, we visualize the prediction results of trajectory for three different maneuvers at different missing rate intervals, as shown in Fig. \ref{fig5}. Compared with the lane-changing trajectory, the lane-keeping trajectory achieves the most accurate prediction results, and the prediction results are insensitive to the missing rate due to its simple behavior. In the case of the left lane changing, the continuity representation extracted by IIPA guides the MSTF to predict a lane-keeping trajectory that is more consistent with the historical motion trend, since the vehicle changes lanes within the prediction time horizon and the historical trajectory does not show a trend of changing lanes. In the case of right lane changing, the model is able to accurately output the right lane-changing trajectory that is consistent with the motion trend when the missing rate is less than 60\%. However, as the missing rate increases to the interval $\left( {60\% ,90\% } \right]$, the MSTF cannot effectively extract detailed motion information, and the model tends to perform lane-keeping prediction. The visualization results show that our model can effectively perform reasonable prediction that consistent with motion consistency for the incomplete trajectory with the missing rate less than 60\%.

\section{CONCLUSIONS AND DISCUSSION}
This paper presents a novel end-to-end framework named MSTF for the incomplete trajectory prediction task, which integrates the Multiscale Attention Head (MAH) and Information Increment-based Adaptive (IIPA) module. We utilize the padding mask matrix in the multi-head attention mechanism to construct the MAH for extracting multiscale motion representations with global dependency from different temporal granularities, so as to alleviate the lack of local dependence caused by random missing values. IIPA analyzes the information increment of different trajectory points through the missing patterns of trajectory, and uses them as weights to aggregate multi-scale representations across time steps to obtain continuity representations. The continuity representation ignores individual missing values and describes the overall trend of motion from a high level so that the MSTF outputs predictions that are consistent with motion consistency.

In the future, we will continue to explore the positive role of HD maps in the task of incomplete trajectory prediction, and further strengthen the prediction performance for incomplete trajectory through the scene constraints extracted from HD maps, so that the model can output predictions that are consistent with scenes in complex traffic scenarios such as those provided by Argoverse.

\bibliography{IEEEref}
\addtolength{\textheight}{-12cm}   






\end{document}